\pdfoutput=1

\documentclass[11pt]{article}
\usepackage{acl}

\usepackage{times}
\usepackage{latexsym}

\usepackage[T1]{fontenc}

\usepackage[utf8]{inputenc}

\usepackage{microtype}

\usepackage{hyperref}
\usepackage{multirow}
\usepackage{graphicx}
\usepackage{tabularx}
\usepackage{url}
\usepackage{xcolor}
\usepackage{epsfig}
\usepackage{adjustbox}
\usepackage{amsfonts, amsmath, amssymb}
\usepackage{booktabs} 
\usepackage{comment}
\usepackage{caption, subcaption}
\usepackage{textcomp}
\usepackage{relsize}
\usepackage{stmaryrd}
\usepackage{bbm}
\usepackage{rotating}
\usepackage{helvet}
\usepackage{courier}
\usepackage{cleveref}
\usepackage{xspace}
\usepackage{enumitem}
\usepackage{mdframed}
\usepackage{pifont}

\newcolumntype{W}{>{\centering\arraybackslash}m{0.06\linewidth}}
\newcolumntype{L}{>{\arraybackslash}m{0.99\linewidth}}

\title{Evidence-Focused Fact Summarization for Knowledge-Augmented Zero-Shot~Question Answering}

\author{Sungho Ko$^{\ast}$, Hyunjin Cho\thanks{\ \ Equal contribution}\ , Hyungjoo Chae, Jinyoung Yeo, Dongha Lee\thanks{\ \ Corresponding author}\\
Yonsei University\\
  \texttt{\{k133324,cyberhyunjin,mapoout,jinyeo,donalee\}@yonsei.ac.kr}}

\newcommand{\gptturbo}{GPT-3.5-turbo\xspace}
\newcommand{\flantfive}{Flan-T5-XL\xspace}
\newcommand{\llamachat}{Llama2-7B-Chat\xspace}

\newcommand{\kaping}{KAPING\xspace}
\newcommand{\rewrite}{Rewrite\xspace}
\newcommand{\kgtotext}{KG2Text\xspace}
\newcommand{\proposed}{\textsc{EFSum}\xspace}
\newcommand{\proposedprompt}{\textsc{EFSum}\textsubscript{$prompt$}\xspace}
\newcommand{\proposeddistill}{\textsc{EFSum}\textsubscript{$distill$}\xspace}

\newcommand{\webqsp}{WebQSP\xspace}
\newcommand{\mintaka}{Mintaka\xspace}

\newcommand{\smallsection}[1]{{\vspace{0.05in} \noindent \bf {#1.\hspace{5pt}}}}

\begin{document}
\maketitle
\begin{abstract}
Recent studies have investigated utilizing Knowledge Graphs (KGs) to enhance Question Answering (QA) performance of Large Language Models (LLMs), yet structured KG verbalization remains challenging. 
Existing methods, such as \textit{triple-form} or \textit{free-form} textual conversion of triple-form facts, encounter several issues. 
These include reduced evidence density due to duplicated entities or relationships, and reduced evidence clarity due to an inability to emphasize crucial evidence. 
To address these issues, we propose {\proposed}, an \textbf{E}vidence-focused \textbf{F}act \textbf{Sum}marization framework for enhanced QA with knowledge-augmented LLMs. 
We optimize an open-source LLM as a fact summarizer through distillation and preference alignment. 
Our extensive experiments show that \proposed improves LLM’s zero-shot QA performance, and it is possible to ensure both the helpfulness and faithfulness of the summary.
\end{abstract}


\section{Introduction}
\label{sec:intro}
Large Language Models (LLMs) have shown remarkable zero-shot abilities but often produce factual errors, known as \textit{hallucinations}, particularly in knowledge-intensive tasks like Question Answering (QA). 
This happens because the static knowledge within LLM parameters may be incomplete, incorrect, or outdated, failing to keep pace with evolving real-world knowledge.
Recent studies remedy this by integrating external knowledge into LLMs~\cite{Karpukhin2020DensePR, Min2019KnowledgeGT}.

As one form of external knowledge, Knowledge Graphs (KGs) have been considered as the knowledge source to augment LLMs for enhanced performance in knowledge graph QA (KGQA)~\cite{baek2023knowledgeaugmented,wu2023retrieve,sen-etal-2023-knowledge}. 
The key challenge of utilizing KGs, which consist of a set of (head entity, relation, tail entity) triples, is to bridge the modality gap between graphs and text. 
Efforts on connecting the gap mostly fall into either training additional layers to blend two representations, or verbalizing graphs into texts. 
While training fusion layers~\cite{yasunaga2022aqagnn, yasunaga2022deep} can make models expressive on two different modalities, it takes expensive computaions, and needs to be trained when KGs are updated. 
On the other hand, recent studies proposed verbalizing KGs into text-form, without training LLMs for QA.
For instance, one strategy is simply concatenating the facts in their \textit{triple-form} text~\cite{baek2023knowledgeaugmented}, and another entails converting them into semantically-coherent textual description (i.e., \textit{free-form} text) through the distillation of LLM's ability to generate text from KGs~\cite{wu2023retrieve}.


\begin{figure}[t]
    \centering
    \includegraphics[width=\linewidth]{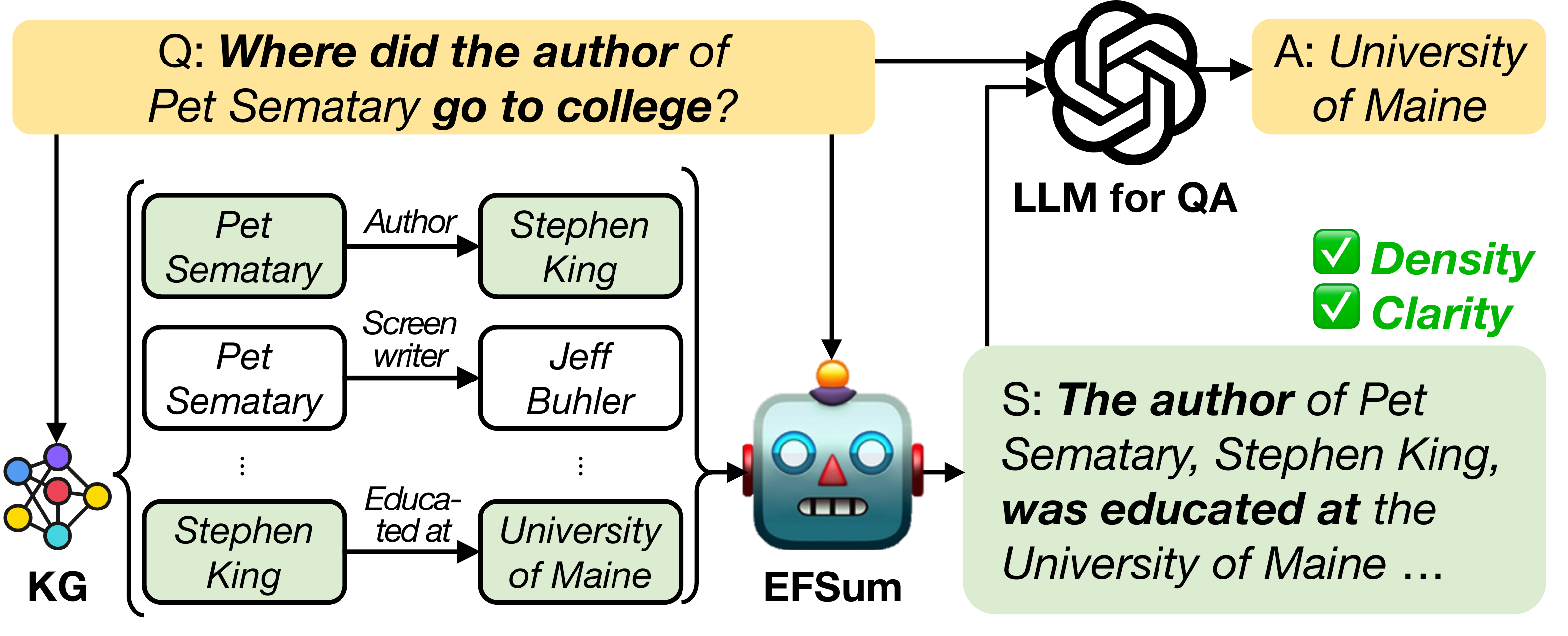}
    \caption{The QA pipeline based on LLM prompting, augmented with relevant facts from KGs. Our fact summarization improves both density and clarity of evidence within contextual knowledge for enhanced QA.}
    \label{fig:pipeline}
\end{figure}

Despite their remarkable efficacy, existing verbalization strategies for providing contextual knowledge exhibit critical limitations.
\textbf{(1) Low density of evidence}: 
Both concatenation and linearization of the facts~\cite{baek2023knowledgeaugmented, wu2023retrieve} are highly likely to include duplicated entities or relations due to limited flexibility;
this eventually degrades the density of useful evidence within the contextual knowledge, hindering the ability to answer questions effectively.
\textbf{(2) Low clarity of evidence}:
While contextual knowledge describes factual information, they often fail to highlight the evidence necessary for answering questions. 
This lack of focus can lead to noise from irrelevant facts, which can detrimentally impact the LLM's ability to provide accurate answers. 

To address the aforementioned challenges, we introduce a novel \textbf{\proposed} framework, \textbf{E}vidence-focused \textbf{F}act \textbf{Sum}marization for enhanced QA with knowledge-augmented LLM prompting. 
The key idea is transforming the set of facts into plausible and coherent \textit{summary} while highlighting evidence and filtering out noise given a question; this ensures that the summaries maintain a high density and clarity of evidence, facilitating effective QA (Figure~\ref{fig:pipeline}).
The most straightforward solution to this summarization is prompting LLMs with the detailed instruction.
However, LLM-generated summaries often omit crucial evidence (such as answer spans), resulting in \textit{information loss}, or include information that cannot be inferred from retrieved facts, leading to \textit{extrinsic hallucination}.

For enhanced summary quality, we optimize an open-source LLM as a fact summarizer in two steps: LLM distillation and preference alignment (Figure~\ref{fig:framework}).
During the first step, we train our fact summarizer by using the reference summaries obtained through LLM prompting.
Subsequently, in the second step,
we refine our summarizer to better align with the task-specific preference related to QA.
To this end, we introduce two preference criteria for the summary candidates:
\textit{helpfulness} evaluates LLMs can correctly answer the question based on the summary, and \textit{faithfulness} assesses the factual consistency of the summary in relation to the provided set of facts.
By selecting pairs of preferred and dispreferred summaries based on these criteria, we further fine-tune the summarizer through direct preference optimization (DPO)~\cite{rafailov2023direct}.
In the end, \proposed is capable of generating summaries that are both helpful for QA and faithful to the given facts.

Extensive experiments on two QA benchmark datasets validate the effectiveness of our evidence-focused fact summarization in improving LLM's zero-shot QA performance. 
\proposed outperforms other fact verbalization methods in two key settings: (1) when fixing the token length (density, Section \ref{subsec:rqone}), and (2) when fixing the number of triples (clarity, Section \ref{subsec:rqtwo}) within contextual knowledge.
Additionally, our approach enhances the helpfulness and faithfulness of the generated fact summaries.
{For reproducibility, our codes are publicly available at \href{https://github.com/kk13332488/EFSum}{https://github.com/kk13332488/EFSum}}.

    
    

\section{Preliminaries}
\label{sec:prelim}

In this section, we introduce a KG-augmented zero-shot QA pipeline, and provide analyses on the verbalized facts obtained by existing methods.

\subsection{KG-Augmented LLM Prompting for QA}
We focus on a QA approach that leverages LLMs' zero-shot capability for answering the question, enhanced with external knowledge from KGs.

\smallsection{Fact retrieval from knowledge graph}
\label{subsec:retrieval}
The first stage aims to retrieve question-associated facts from KGs via entity linking, and then to select only top-$K$ ones based on their semantic relevance to the input question.
To select only the most relevant facts, recent studies utilize semantic similarities between each fact and the question, employing either a pretrained sentence encoder~\cite{karpukhin-etal-2020-dense, xiong2020approximate} or one fine-tuned specifically for direct fact retrieval~\cite{baek-etal-2023-direct}.
In this work, we utilized the former strategy~\cite{song2020mpnet}, if not stated.


\smallsection{Fact verbalization into various form text}
\label{subsec:verbalization}
The second stage is fact verbalization, which refers to the task of transforming symbolic facts into textual strings, for feeding them into the LLM as the contextual knowledge.
The linear verbalization simply concatenates the head, relation, and tail texts in the triple while keeping the structured format (i.e., triple-form text)~\cite{baek2023knowledgeaugmented}, or use manually-designed templates and heuristics for linearization~\cite{oguz-etal-2022-unik, ma2022open}.
On the other hand, the graph-to-text verbalization transforms the input facts into the plausible and coherent text by using a fine-tuned model~\cite{ribeiro-etal-2021-investigating} or prompting LLMs~\cite{wu2023retrieve}.

\smallsection{Fact injection for question answering}
\label{susbec:llmqa}
The last stage is prompting the LLM to generate the answer with the verbalized facts.
This process, also known as knowledge-augmented LLM prompting for zero-shot QA,
gathers the output as the predicted answer.
To handle with insufficient evidence of knowledge, the detailed instructions are provided to allow the LLM to utilize its internal knowledge if needed.
The prompts are in Appendix~\ref{subsec:detailedprompt}.

\begin{figure}[t]
    \centering
    \includegraphics[width=\linewidth]{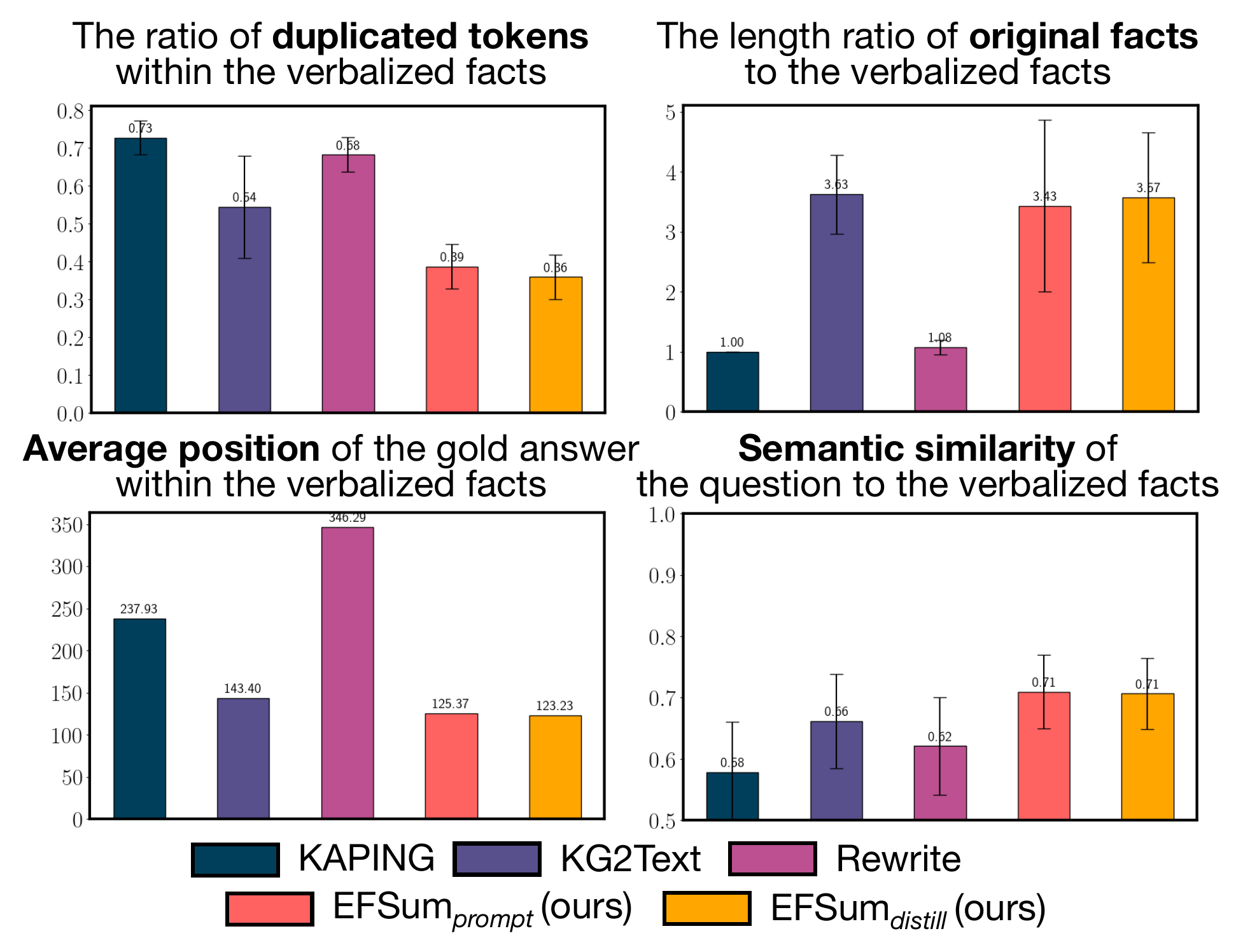}
    \caption{Analysis on each fact verbalization method.}
    \label{fig:verbalization}
\end{figure}

\subsection{Analysis on Verbalized Facts}
We provide a preliminary analysis of the contextual knowledge obtained by each fact verbalization method, evaluating their (1) density and (2) clarity of evidence.
Note that our proposed verbalization method (denoted as \proposed) summarizes the symbolic facts into free-form texts, prioritizing the evidence relevant to the question.

In Figure~\ref{fig:verbalization} Upper, it is evident that the number of duplicated tokens is significantly higher in the linearly verbalized texts (i.e., \kaping\cite{baek2023knowledgeaugmented} and \rewrite\cite{wu2023retrieve}) compared to the others. 
This strongly indicates that their outputs contain redundant information stemming from the pre-defined relations within KGs.
When considering the ratio of token lengths before and after verbalization (i.e., compression rate), \kaping and \rewrite either maintain or even increase the length, despite maintaining the same amount of information. 
Consequently, their evidence density remains low or may even decrease during verbalization.

In Figure~\ref{fig:verbalization} Lower, the linear verbalization tends to scatter obvious evidence (i.e., answer span) randomly within the contextual knowledge. 
Their placement seems to rely on the rank obtained from fact retrieval, rather than being positioned at the forefront of the verbalized texts for emphasis.
Moreover, the average semantic similarity \footnote{Average semantic similarity is calculated through the cosine similarity between embeddings encoded via MPNet.} of their verbalized facts with the question is lower compared to other methods. 
This shows that their outputs may include noisy or irrelevant information, or they might not clearly highlight semantically relevant evidence.
In essence, clarity of evidence is not adequately addressed in this approach.

\section{\proposed: Proposed Method}
\label{sec:method}


\begin{figure*}[t]
    \centering
    \includegraphics[width=0.95\linewidth]{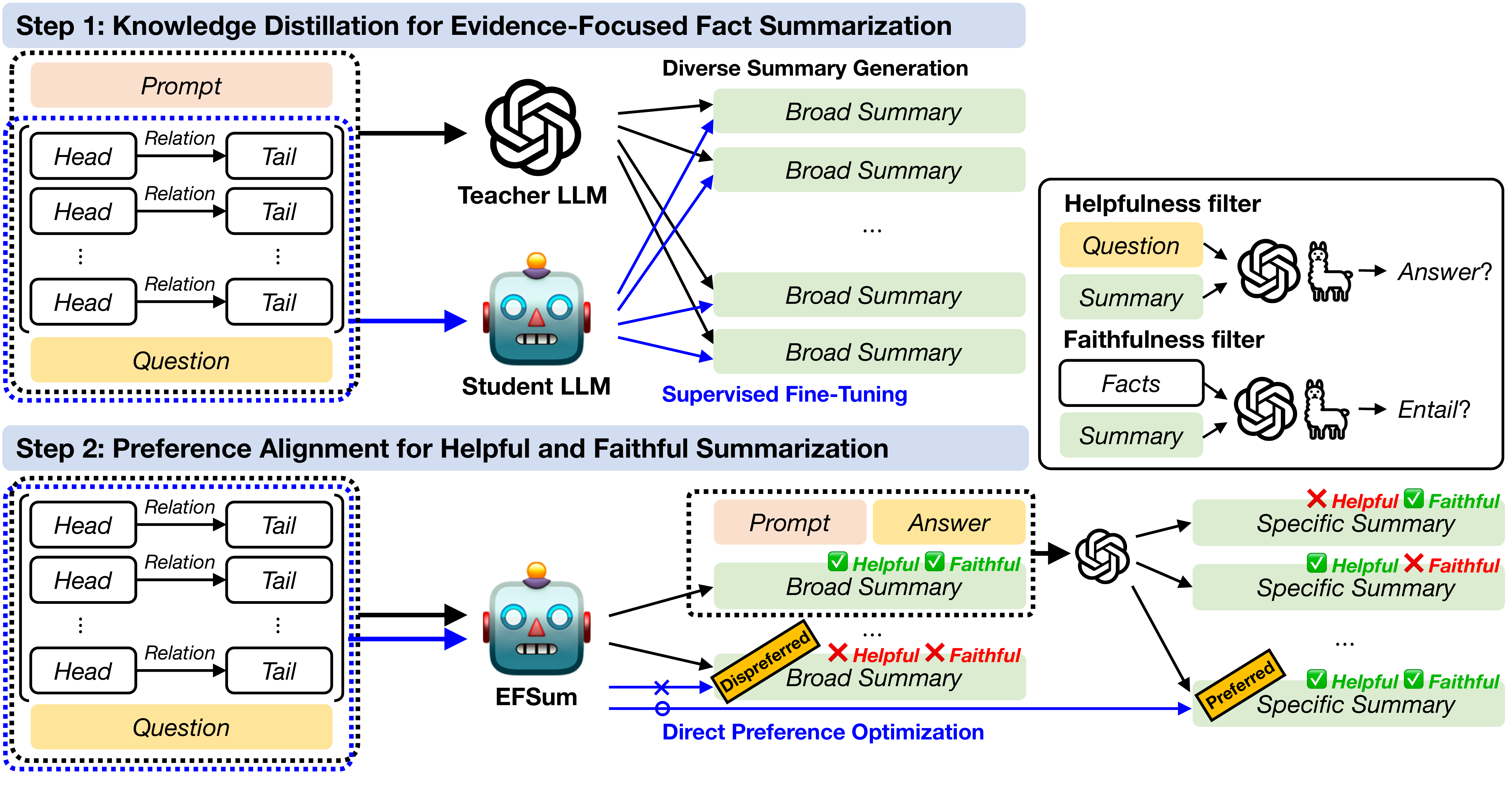}
    \caption{The overall framework of \proposeddistill.
    Our fact summarizer is trained to generate evidence-focused summaries via LLM distillation, and then further optimized to align the QA-specific preference, which enhances the helpfulness and faithfulness of its output summaries.}
    \label{fig:framework}
\end{figure*}

In this section, we present an \textbf{E}vidence-focused \textbf{F}act \textbf{Sum}marization framework, named \textbf{\proposed}, which aims to effectively augment the set of relevant facts to LLMs for zero-shot QA tasks.

\subsection{LLM Prompting for \proposed}
\label{subsec:efsumprompt}

The most straightforward implementation of evidence-focused fact summarization is prompting LLMs to generate a summary $s$ given a question $q$ and its relevant facts $\mathcal{F}=\{f_k\}_{k=1}^K$, focusing on the evidence to answer the question.
Thus, we present \textbf{\proposedprompt} to verbalize the facts with the help of LLM's zero-shot capability on summarization;
that is, $s \sim p_{\text{LLM}}(\cdot|t_{sum},q,\mathcal{F})$, where $t_{sum}$ is the prompt for summarization.
Specifically, we instruct LLMs to turn the input facts into the summary for the scenarios where the {summary} serves as a context to facilitate QA task. 
We utilized GPT-3.5-turbo for generating summary of \proposedprompt.
The detailed prompts are in Appendix~\ref{subsec:detailedprompt}.




\subsection{LLM Fine-Tuning for \proposed}
\label{subsec:prompt}
To enhance the quality of summaries, we propose \textbf{\proposeddistill}, a fact verbalization model based on an open-source LLM,\footnote{In this work, we choose Llama2-7B~\cite{touvron2023llama} as the backbone open-source LLM for \proposeddistill.} specifically fine-tuned for evidence-focused fact summarization.
For helpful and faithful summary generation, \proposeddistill is optimized in two steps: (1) LLM distillation and (2) preference alignment.
Figure~\ref{fig:framework} illustrates the overall \proposeddistill framework.

\subsubsection{Distillation of Fact Summarization}
\label{subsubsec:distillation}
We first optimize our open-source model to generate diverse evidence-focused summaries given the question and the set of symbolic facts.
For optimization, we augment the QA training dataset of question-answer pairs $(q, a)$ into $\mathcal{X}=\{(q, a, \mathcal{F})\}$, where $\mathcal{F}=\{f_k\}_{k=1}^K$ is the top-$K$ retrieved facts relevant to the question $q$.

\smallsection{Reference summary generation}
We utilize a closed-source LLM (i.e., GPT-3.5-turbo) to obtain reference fact summaries used for training our summarizer.
For each tuple $(q, a, \mathcal{F})$ of a question $q$ and the relevant facts $\mathcal{F}$, we prompt the LLM to transform the set of facts $\mathcal{F}$ into a concise textual description $s$ that highlights the evidence for the question $q$, which is same with \proposedprompt.
In the end, we construct the training dataset $\mathcal{D}=\{(q,a,\mathcal{F},s)\}$. The prompt used for summary generation is in Table 5 of Appendix~\ref{subsec:detailedprompt}.

\smallsection{Supervised fine-tuning}
For each quadruplet $(q,\mathcal{F},s, a)$ from the dataset $\mathcal{D}$, our summarizer $\theta$ is optimized to generate $s$ conditioned on $q$ and $\mathcal{F}$, by using the causal language modeling objective:
\begin{equation}
    \mathcal{L}_{\text{SFT}} = -\underset{(q,a,\mathcal{F},s)\sim D}{\mathbb{E}} \log p_\theta(s|q, \mathcal{F}).
\end{equation} 

\subsubsection{Alignment with Summary Preference}
\label{subsec:preferalignment}

Our summarizer $\theta$ is now able to generate evidence-focused summaries for a question and the relevant facts, but its output summaries might be unhelpful or unfaithful.
Therefore, we additionally adopt preference tuning to enhance \proposeddistill so that its summarization can align with the task-specific preference in the context of knowledge-augmented zero-shot QA;
i.e., generating helpful and faithful summaries, while avoiding the counterparts.


For preference tuning, we collect a set of preference pairs $(q,a,\mathcal{F},s^+,s^-)$, where $s^+$ and $s^-$ respectively denote the \textit{preferred} and \textit{dispreferred} summaries.
To identify the preferred and dispreferred summary for QA task, we first sample $M$ summary candidates $\{s'_m \sim p_{\theta}(\cdot|q, \mathcal{F}) \}_{m=1}^M $ for each tuple $(q,a,\mathcal{F},s)\in\mathcal{D}$ using the summarizer $\theta$.
Then, for constructing the preference pairs, we adopt two summary filters (for checking helpfulness and faithfulness) and additional paraphrase process.
We form a preference pair by selecting the answer-aware paraphrased candidate that passes both filter as {preferred}, and the candidate that cannot pass one of the filters as {dispreferred}.


\smallsection{Helpfulness filter}
\label{subsec:stepthree}
The first filter examines the helpfulness of each summary candidate in terms of QA accuracy.
That is, it checks whether the summary candidate $s'$ is actually helpful to make the LLM to find the correct answer, by comparing the LLM's generated answer $a'\sim p_{\text{LLM}}(\cdot|q, s')$ with the gold answer $a$.


\smallsection{Faithfulness filter}
\label{subsec:stepthree}
The second filter focuses on the faithfulness of each summary candidate.
We use the G-Eval approach~\cite{liu-etal-2023-g} to leverage LLM's ability to evaluate the consistency between the input facts and the given summary in terms of hallucination.
Precisely, the LLM is prompted to examine whether or not the summary contains unfaithful information, which cannot be inferred from the given symbolic facts.
Please refer to the detailed prompt in Appendix~\ref{subsec:detailedprompt}.

\smallsection{Broad-to-specific paraphrasing}
In addition, we adopt the paraphrase process guided by the gold answer $a$.
This paraphrasing aims to obtain high-quality summary by refining a broad focus into a specific focus, using the given answer as the main evidence.
We prompt the LLM to paraphrase the summary candidate $s'$ into $s''$;  $\{s''_m \sim p_{\text{LLM}}(\cdot|t_{\text{paraphrase}},s',a)\}_{m=1}^M$, where $t_{\text{paraphrase}}$ is the prompt for paraphrasing.
This candidate \(s''\) undergoes another pass through the helpfulness filter and the faithfulness filter. The resulting summary obtained in this manner is much more focused on the QA task than the initially obtained reference summary, called as the broad summary. 
The detailed prompt is provided in Appendix~\ref{subsec:detailedprompt}

\smallsection{Direct Preference Optimization}
Using the preference pairs $\mathcal{P}=\{(q,a, \mathcal{F},s^+,s^-)\}$, we apply Direct Preference Optimization (DPO)~\cite{rafailov2023direct} on our summarizer $\theta$ to train a preference-tuned summarizer $\theta^*$ that minimizes the following objective:
\begin{equation}
\begin{split}
    &\mathcal{L}_{\text{DPO}}(\theta^*; \theta) = \\
    &-\underset{(q,a,\mathcal{F},s^+,s^-)\sim P}{\mathbb{E}} \log \sigma[r(q, \mathcal{F},s^+)-r(q,\mathcal{F},s^-)],
\end{split}
\raisetag{40pt}
\end{equation}
where $r(q,\mathcal{F},s)=\frac{p_{\theta^*}(s|q,\mathcal{F})}{p_\theta(s|q,\mathcal{F})}$.
By optimizing the model using preferred-dispreferred summary pairs, our obtained model $\theta^*$ is trained to be more biased towards helpful and faithful summary $s^+$ while avoiding unhelpful or unfaithful summary $s^-$.
Note that $\theta$ has been specifically trained for each QA model, as there are different preferences regarding summary helpfulness across various QA models.
The following is an example of an output fact summary.

\begin{mdframed}
\label{tbl:casestudy}
\footnotesize
\noindent\textbf{Question:}
where was george washington carver from?

\noindent\textbf{Answer:}
Diamond

\noindent\textbf{Facts:}
(George Washington Carver, occupation, biologist),(George Washington Carver, interested in, botany),(George Washington Carver, occupation, university teacher),(George Washington Carver, place of birth, Diamond),(George Washington Carver, given name, George), (George Washington Carver, field of work, pedagogy),(George Washington Carver, relative, Moses Carver),(George Washington Carver, place of birth, United States of America),(George Washington Carver, residence, Tuskegee),(George Washington Carver, occupation, inventor)

\noindent\textbf{Summary:}
George Washington Carver, the renowned biologist, was born in Diamond, United States of America. He took great interest in botany and became a university teacher, focusing on pedagogy. Additionally, Carver contributed significantly as an inventor. His relative, Moses Carver, also had a close association with him. Carver later resided in Tuskegee.

\end{mdframed}

\section{Experiments}
\label{sec:exp}

\begin{table*}[t]
\centering
\small
\begin{tabular}{clWWWWWW}
\toprule
                         \multirow{2.5}{*}{\textbf{Dataset}} & \multirow{2.5}{*}{\textbf{Methods}} & \multicolumn{2}{c}{\textbf{\gptturbo}}                  & \multicolumn{2}{c}{\textbf{\flantfive}}                    & \multicolumn{2}{c}{\textbf{\llamachat}}                     \\ \cmidrule(lr){3-8} 
                         & & $L$=200 & $L$=400 & $L$=200 & $L$=400 & $L$=200 & $L$=400 \\ \midrule \midrule
\multirow{7}{*}{\textbf{\webqsp}}  & No knowledge & 0.506 & 0.506 & 0.409 & 0.409 & {0.539} & {0.539} \\ \cmidrule{2-8} 
                         & \kaping~\cite{baek2023knowledgeaugmented} & 0.507 & \underline{0.538} & 0.391 & 0.439 & \textbf{0.517} & \textbf{0.519} \\
                         & \kgtotext~\cite{ribeiro-etal-2021-investigating} & 0.476 & 0.476 & 0.316 & 0.321 & 0.439 & 0.481 \\
                         & \rewrite~\cite{wu2023retrieve} & 0.444 & 0.525 & 0.350 & 0.431 & 0.462 & \underline{0.511} \\
                         & \proposedprompt (Ours)  & \underline{0.537} & \underline{0.538} & \underline{0.447} & \underline{0.468} & 0.457 & 0.491 \\
                         & \proposeddistill (Ours) & \textbf{0.559} & \textbf{0.569} & \textbf{0.458} & \textbf{0.500} & \underline{0.489} & 0.497 \\ \midrule

\multirow{7}{*}{\textbf{\mintaka}}  & No knowledge & 0.540 & 0.540 & 0.228 & 0.228 & 0.440 & 0.440 \\ \cmidrule{2-8} 
                         & \kaping~\cite{baek2023knowledgeaugmented} & \textbf{0.539} & \textbf{0.539} & 0.269 & 0.279 & 0.402 & \underline{0.407} \\
                         & \kgtotext~\cite{ribeiro-etal-2021-investigating} & 0.492 & 0.491 & 0.234 & 0.234 & 0.377 & 0.378 \\
                         & \rewrite~\cite{wu2023retrieve} & \underline{0.515} & \underline{0.521} & 0.280 & 0.288 & 0.394 & 0.386 \\
                         & \proposedprompt (Ours)  & 0.496 & 0.491 & \underline{0.312} & \underline{0.321} & \textbf{0.423} & \textbf{0.418} \\
                         & \proposeddistill (Ours) & 0.474 & 0.449 & \textbf{0.326} & \textbf{0.338} & \underline{0.405} & 0.406  \\ \bottomrule
\end{tabular}
\caption{QA accuracy of the LLMs based on various fact verbalization. We limit the maximum token length of contextual knowledge to $L=200$ and $400$. The best and second-best results are in \textbf{bold} and \underline{underlined}, respectively.}
\label{tbl:main_density}
\end{table*}

\begin{table*}[t]
\centering
\resizebox{.99\textwidth}{!}{
\begin{tabular}{clWWWWWWWWW}
\toprule
\multirow{2.5}{*}{\textbf{Datasets}} & \multirow{2.5}{*}{\textbf{Methods}}  & \multicolumn{3}{c}{\textbf{\gptturbo}} & \multicolumn{3}{c}{\textbf{\flantfive}} & \multicolumn{3}{c}{\textbf{\llamachat}} \\ \cmidrule(lr){3-11} 
& & {\small Random}    & {\small Popular}    & {\small MPNet}    & {\small Random}   & {\small Popular}   & {\small MPNet}   & {\small Random}    & {\small Popular}    &  {\small MPNet}    \\ \midrule\midrule

\multirow{7}{*}{\textbf{\webqsp}}
& No knowledge & 0.506 & 0.506 & 0.506 & {0.409} & 0.409 & 0.409 & {0.539} & 0.539 & 0.539 \\ \cmidrule{2-11}
& \kaping~\cite{baek2023knowledgeaugmented} & 0.441 & 0.437 & \underline{0.538} & 0.297 & 0.329 & 0.439 & \underline{0.476} & \textbf{0.490} & \textbf{0.519} \\
& \kgtotext~\cite{ribeiro-etal-2021-investigating} & 0.469 & 0.468 & 0.476 & 0.317 & 0.276 & 0.321 & 0.465 & 0.451 & 0.481  \\
& \rewrite~\cite{wu2023retrieve} & {0.473} & 0.445 & 0.525 & 0.323 & 0.348 & 0.431 & 0.458 & 0.439 & \underline{0.511} \\
& \proposedprompt (Ours) & \textbf{0.542} & \underline{0.534} & \underline{0.538} & \underline{0.443} & \underline{0.442} & \underline{0.468} & \textbf{0.477} & 0.472 & 0.491 \\
& \proposeddistill (Ours) & \underline{0.475} & \textbf{0.539} & \textbf{0.569} & \textbf{0.500} & \textbf{0.505} & \textbf{0.500} & 0.457 & \underline{0.488} & 0.497  \\ \midrule

\multirow{7}{*}{\textbf{\mintaka}}
& No knowledge & 0.540 & 0.540 & 0.540 & 0.228 & 0.228 & 0.228 & 0.440 & 0.440 & 0.440 \\ \cmidrule{2-11}
& \kaping~\cite{baek2023knowledgeaugmented} & \textbf{0.553} & \underline{0.516} & \textbf{0.539} & 0.201 & 0.198 & 0.279 & \underline{0.417} & \textbf{0.398} & \underline{0.407} \\
& \kgtotext~\cite{ribeiro-etal-2021-investigating} & 0.505 & 0.500 & 0.492 & 0.220 & \underline{0.235} & 0.234 & \textbf{0.421} & 0.389 & 0.378 \\
& \rewrite~\cite{wu2023retrieve} & \underline{0.527} & \textbf{0.524} & \underline{0.515} & \underline{0.230} & 0.224 & 0.288 & 0.393 & 0.374 & 0.386 \\
& \proposedprompt (Ours) & 0.454 & 0.492 &  0.496 & 0.213 & 0.215 & \underline{0.321} & 0.390 & 0.392 & \textbf{0.418} \\
& \proposeddistill (Ours) & 0.427 & 0.425 & 0.474 & \textbf{0.292} & \textbf{0.243} & \textbf{0.338} & 0.397 & \underline{0.393} & {0.406} \\
\bottomrule

\end{tabular}
}
\caption{QA accuracy of the LLMs based on various fact verbalization, with different fact retrieval strategies (i.e., random facts, popular facts, and question-relevant facts). We limit the maximum token length of contextual knowledge to $L=400$. The best and second-best results are in \textbf{bold} and \underline{underlined}, respectively.}
\label{tbl:retriever}
\end{table*}

In this section, we design our experiments to answer the following research questions:
\begin{itemize}
[leftmargin=*,topsep=2pt,itemsep=2pt,parsep=0pt]
\item \textbf{RQ1:} Does a high density of evidence in verbalized facts contribute to QA accuracy?
\item \textbf{RQ2:} Does a high clarity of evidence in verbalized facts contribute to QA accuracy?
\item \textbf{RQ3:} Can preference alignment enhance the generation of more helpful and faithful summaries?
\end{itemize}

\subsection{Experimental Settings}
\label{subsec:expset}

\smallsection{Datasets}
\textbf{WebQuestionsSP (WebQSP)}~\cite{yih-etal-2016-value} is a KGQA dataset that filters questions from the WebQuestions~\cite{berant-etal-2013-semantic} dataset to include only those answerable via Freebase, and provides SPARQL queries for them.
For convenience, we use WebQSP-WD~\cite{sorokin-gurevych-2018-modeling}, in which each question from WebQSP is pre-linked to the Wikidata KG.
We use a test set comprising 1,033 examples for evaluation. 
\textbf{Mintaka}~\cite{sen-etal-2022-mintaka} is a QA dataset that encompasses eight different complexity types. 
Most question-answer pairs can only be solved by utilizing multi-hop reasoning or the attributes of multiple entities. 
We use a test set of Mintaka which has 4,000 examples for evaluation. 


\smallsection{LLMs for zero-shot QA}
To measure the efficacy of \proposed and other fact verbalization methods, we utilized three different LLMs, \textbf{\gptturbo}, \textbf{\flantfive}~\cite{chung2022scaling}, and \textbf{\llamachat}~\cite{touvron2023llama}, for our zero-shot QA evaluation.
Note that \flantfive and \llamachat are publicly available as open-source.
We provide more details on the models in Appendix~\ref{subsec:llmdetail}.



\smallsection{Baseline methods}
As the main baselines for fact verbalization, we consider various approaches.
\begin{itemize}
[leftmargin=*,topsep=2pt,itemsep=2pt,parsep=0pt]
\item \textbf{No knowledge} does not pass any knowledge contexts, encouraging the LLMs to use their internal knowledge to answer the question.
\item \textbf{\kaping}~\cite{baek2023knowledgeaugmented} simply linearizes top-$K$ relevant facts as the triple-form text. 
Triple-form text simply refers to the text that is composed by concatenating triplet strings in the form of (head, relation, tail).
\item \textbf{\rewrite}~\cite{wu2023retrieve} transforms facts into the free-form text for each relation path with a LLM. We utilize GPT-3.5-turbo to convert triples into free-form text.
\item \textbf{\kgtotext}~\cite{ribeiro-etal-2021-investigating} employs an encoder-decoder model fined-tuned for the KG-to-text task by using WebNLG~\cite{gardent-etal-2017-webnlg} dataset. We utilize a fine-tuned T5-large model\footnote{https://public.ukp.informatik.tu-darmstadt.de/ribeiro/graph2text/webnlg-t5-large.ckpt} as our base KG2Text model.
\end{itemize}

\smallsection{Evaluation metrics}
\label{subsec:evalmetric}
Our task can be categorized as a generative KGQA. 
Following previous work ~\cite{baek2023knowledgeaugmented, wu2023retrieve}, we use accuracy as our evaluation metric. 
A score of 1 is assigned if at least one among multiple correct answers is present in the response text of the QA model; otherwise, the score is 0.

\smallsection{Relevant fact retrieval}
\label{subsec:implementation}
In retrieving question-related facts, as described in the Section \ref{subsec:retrieval}, we employ MPNet \cite{song2020mpnet} to retrieve only the top-$K$ triples among given KG, with the highest semantic similarity to the question representation, following \citet{baek2023knowledgeaugmented}. 
Since processing entire KGs is impractical, we focus on retrieving information from the $n$-hop neighbors of question entities within a KG. The value of $n$ is given by the specific KGQA dataset to answer the question. In our experiment, $n$ is set to 1 for WebQSP and 2 for Mintaka.
When calculating semantic similarity, we use the linear verbalization approach, which involves combining the subject, relation, and object texts from the triple.

\begin{table*}[thbp]
\centering
\small
\begin{tabular}{clWWWWWW}
\toprule
                         \multirow{2.5}{*}{\textbf{Dataset}} & \multirow{2.5}{*}{\textbf{Methods}} & \multicolumn{2}{c}{\textbf{\gptturbo}}                  & \multicolumn{2}{c}{\textbf{\flantfive}}                    & \multicolumn{2}{c}{\textbf{\llamachat}}                     \\ \cmidrule(lr){3-8} 
                         & & $K$=10 & $K$=30 & $K$=10 & $K$=30 & $K$=10 & $K$=30 \\ \midrule \midrule
\multirow{7}{*}{{\textbf{\webqsp}}}  & No knowledge & 0.617 & 0.607 & {{0.498}} & {{0.451}} & {0.646} & 0.628 \\ \cmidrule{2-8} 
                         & \kaping~\cite{baek2023knowledgeaugmented} & 0.777 & \underline{0.771} & \underline{0.643} & \underline{0.738} & \textbf{0.668} & \textbf{0.699} \\
                         & \kgtotext~\cite{ribeiro-etal-2021-investigating} & 0.589 & 0.608 & 0.467 & 0.457 & 0.409 & 0.536 \\
                         & \rewrite~\cite{wu2023retrieve} & 0.628 & 0.728 & 0.533 & 0.664 & 0.594 & \underline{0.688} \\
                         & \proposedprompt (Ours)  & \textbf{0.788} & 0.755 & 0.629 & 0.711 & 0.497 & 0.571 \\
                         & \proposeddistill (Ours) & \underline{0.786} & \textbf{0.783} & \textbf{0.644} & \textbf{0.741} & \underline{0.599} & 0.666 \\ \midrule

\multirow{7}{*}{{\textbf{\mintaka}}}  & No knowledge  & 0.810 & 0.788 & 0.492 & 0.444 & 0.783 & 0.719 \\ \cmidrule{2-8} 
                         & \kaping~\cite{baek2023knowledgeaugmented} & \underline{0.912} & 0.869 & 0.723 & 0.673 & 0.832 & \underline{0.808} \\
                         & \kgtotext~\cite{ribeiro-etal-2021-investigating} & 0.879 & 0.768 & 0.536 & 0.491 & 0.799 & 0.727 \\
                         & \rewrite~\cite{wu2023retrieve} & 0.901 & \underline{0.875} & 0.720 & 0.691 & 0.843 & 0.792 \\
                         & \proposedprompt (Ours)  & \textbf{0.920} & \textbf{0.887} & \underline{0.742} & \textbf{0.735} & \underline{0.849} & 0.806 \\
                         & \proposeddistill (Ours) & 0.893 & 0.824 & \textbf{0.745} & \underline{0.719} & \textbf{0.852} & \textbf{0.826} \\ \bottomrule
\end{tabular}
\caption{QA accuracy of the LLMs based on fact verbalization methods. We fix the number of facts to $K=10$ and $30$. The best and second-best results are in \textbf{bold} and \underline{underlined}, respectively.}
\label{tbl:main_clarity}
\end{table*}

\subsection{Effectiveness of Dense Evidence (RQ1)}
\label{subsec:rqone}
To examine the impact of dense evidence within verbalized facts on the final QA performance, we evaluate the LLM's QA accuracy while imposing restrictions on the maximum token lengths $L$ of contextual knowledge.
This implies that the number of facts included in the contextual knowledge varies depending on fact verbalization methods.

\smallsection{Effect of knowledge augmentation}
First of all, we observe that knowledge augmentation for zero-shot QA does not always produce positive results.
Knowledge augmentation cannot be helpful in two scenarios: where the model's ability to ground input knowledge is lacking, or where the retrieved knowledge is noisy while the QA model's internal knowledge is sufficient. 
In Table \ref{tbl:main_density}, when \llamachat was used as the QA model, it demonstrates higher performance under the ``No knowledge'' condition across both datasets (i.e., \webqsp and \mintaka) compared to other baselines. 
This is indicative of \llamachat's limited capability in utilizing the provided knowledge. 

\smallsection{Comparison with other baselines}
We first compare the performance of various fact verbalization methods at $L=200$ and $400$.
In Table~\ref{tbl:main_density}, for most cases, \proposed shows superior performance over the majority of baseline approaches. 
Considering the ratio of lengths before/after verbalization (in Figure~\ref{fig:verbalization}), this clearly indicates that \proposed can encapsulate more intensive and useful information within shorter summaries. 
Notably, the effectiveness of our approach is more evident when the length of knowledge decreases from $L=400$ to $200$.
This suggests that \proposed remains highly effective even in contexts when the utilization of extremely concise knowledge is required. 

\smallsection{Compatibility with various retrievers}
We investigate the QA performance using different types of fact retrieval, to assess the robustness across various knowledge qualities: randomly selected knowledge (\textbf{Random}), the knowledge possessing the most frequently occurring relation (\textbf{Popular}), and question-relevant knowledge (\textbf{MPNet}). 
In Table~\ref{tbl:retriever}, \proposed achieves the highest accuracy across most datasets and QA models, regardless of the retriever used, which implies that our method can extract useful facts from noisy input triples.

\begin{figure}[t]
    \centering
    \includegraphics[width=\linewidth]{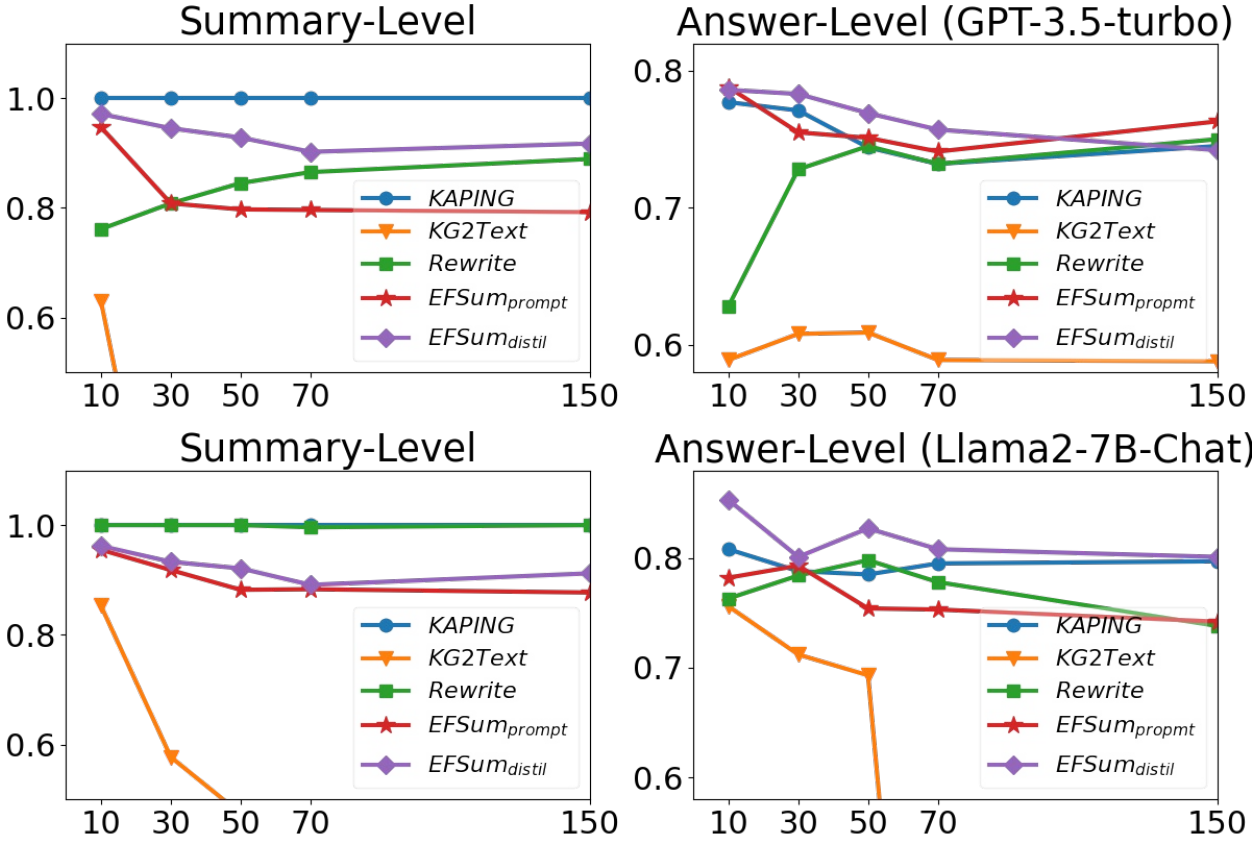}
    \caption{Summary-level and answer-level QA accuracies with respect to the number of relevant facts 
    on {WebQSP} (Upper) and {Mintaka} (Lower), respectively.
    }
    \label{fig:robustness}
\end{figure}

\subsection{Effectiveness of Clear Evidence (RQ2)}
\label{subsec:rqtwo}

To examine the effectiveness of clear evidence within verbalized facts on the final QA performance, we assess the LLM's QA accuracy only on the test tuples $(q, a, \mathcal{F})$ where the facts $\mathcal{F}$ fully contain the ground-truth answer span $a$. 
This experimental setup allows us to investigate how effectively each fact verbalization method converts $\mathcal{F}$ into a textual string without overlooking evidence $a$, thereby enabling correct answers.

\smallsection{Comparison with other baselines}
To examine the effectiveness of \proposed in terms of evidence clarity, we evaluate the LLM's zero-shot QA accuracy when the number of facts is $K=10$ and $30$.
In Table~\ref{tbl:main_clarity}, 
\proposed shows the highest performance across most datasets, showing consistently leading results for both $K$ values.
Nevertheless, we notice the presence of uncontrolled model inclination.
For example, \kaping consistently exhibits the highest performance when \llamachat is utilized as the QA model for the WebQSP dataset.
This might be due to the model's inclination towards a specific knowledge format.


\input{043IncreasingK}
\smallsection{Robustness of \proposed across various $K$}
To assess the robustness of \proposed, we increase the retrieved number of facts $K$ from 10 to 150 and measure the performance of each baseline at the \textit{answer-level} and \textit{summary-level}. 
At the \textit{answer-level}, we evaluate the extent to which the answers are contained within the responses generated by the QA model using summaries. In case of the \textit{summary-level}, we evaluate whether the answer is included within the verbalized knowledge produced by the method.
In Figure \ref{fig:robustness}, across both datasets and the majority of \textit{K} values, \proposeddistill consistently outperforms, especially in answer-level accuracy, except for KAPING, which inherently achieves a summary level accuracy of 1 by simple linearization of facts.

\subsection{Effect of Preference Alignment (RQ3)}
\label{subsec:rqthree}

\begin{figure}[t]
    \centering
    \includegraphics[width=\linewidth]{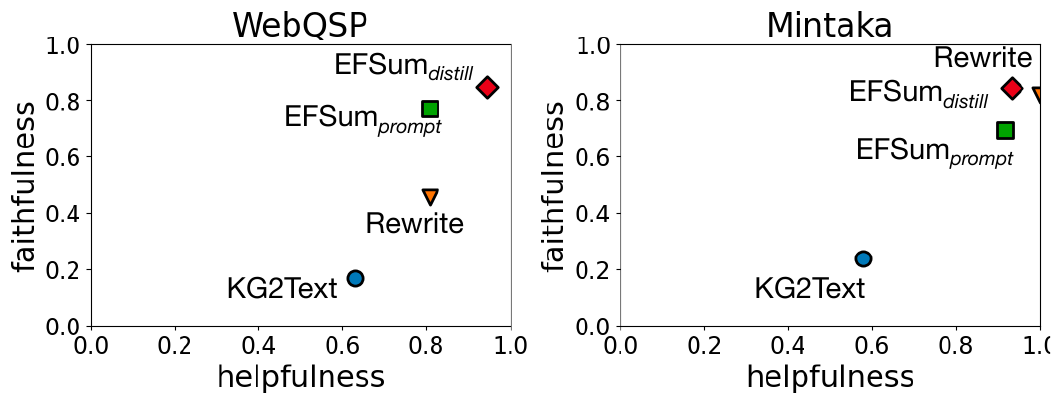}
    \caption{Two quality metrics of verbalized facts.}
    \label{fig:tradeoff}
\end{figure}

\smallsection{Helpfulness and faithfulness}
We examine the \textit{helpfulness} and \textit{faithfulness} of the verbalized facts generated by each verbalization method.\footnote{We opt not to include \kaping in the plot, as its linear verbalization does not alter the content of the facts.}
Note that helpfulness is calculated through summary-level accuracy, while
faithfulness is calculated as 1 - (hallucination occurrence rate). 
That is, a method with a lower rate of hallucination occurrences possesses higher faithfulness.
Figure~\ref{fig:tradeoff} illustrates that \proposeddistill achieves superior helpfulness and faithfulness compared to other baselines, including \proposedprompt.
In other words, \proposeddistill is less likely to generate summaries of hallucination and is capable of incorporating more correct answers into the summaries compared to other baselines.
Moreover, it validates that using the two filters in the broad-to-specific paraphrasing process definitively aids in improving both faithfulness and helpfulness.


\section{Related Work}
\label{sec:relwork}
\subsection{KG-Augmented LLM Prompting}
To supplement the incomplete internal knowledge of LLMs, recent research has been exploring knowledge augmentation methods using prompting. 
Various types of information can be utilized for prompting knowledge, notably KGs. 
Recently, \citet{baek2023knowledgeaugmented, wu2023retrieve} utilize triple-form facts from KGs for knowledge augmentation for LLM-prompting.
However, the knowledge injected into LLMs by these methods is not focused on the question. 
In other words, the semantics of the question are not considered during the process of deriving the final knowledge from the given triples. 
Consequently, the final knowledge derived through each method may inevitably have low density and clarity from an evidence perspective.

\subsection{Knowledge Graph Question Answering}
\label{subsec:kgqa}
Studies exploring the integration of additional knowledge extracted from KGs, represented as subgraphs, are divided into two main approaches. 
The first approach utilizes semantic parsing techniques~\cite{bao-etal-2016-constraint, luo-etal-2018-knowledge}, which enable the extraction of executable queries from the KG by using contextual information as a parsing reference. Alternatively, the information retrieval approach involves encoding assimilated information using techniques like Graph Neural Networks (GNNs).
Several recent studies \cite{yasunaga2022aqagnn, yasunaga2022deep, zhang2022greaselm} propose a learning framework that combines GNNs and LMs, allowing concurrent utilization of textual data and KG.
Besides, approaches that project text embeddings to graph embeddings \cite{razzhigaev-etal-2023-system} struggle to avoid hallucinations by LLMs.
In contrast to the conventional KGQA approaches that aim to directly identify the answer entity within KGs, in this work, we mainly focus on how the KG-retrieved facts can be effectively utilized within the zero-shot QA capability of LLMs.



\section{Conclusion}
\label{sec:conclusion}

In this paper, we explore methods to enhance the zero-shot QA performance of LLMs by augmenting knowledge from KGs. 
We introduce a novel summarization framework, called {\proposed}, which transforms a set of facts into summary with high density and clarity of evidence for answering the question. 
To achieve this, we optimize an open-source LLM as a fact summarizer, leveraging a teacher LLM's summarization capabilities and aligning its outputs with QA-specific preferences.
Our experiments show that {\proposed} significantly improve QA accuracy across various LLMs compared to other fact verbalization approaches. 
Furthermore, serving as an independent summarization module, it generates helpful and faithful summaries based on relevant facts and target questions.

\section{Limitation}
\label{sec:limitation}
Despite our discoveries and improvements, we must acknowledge certain limitations in our work and potential areas for future research.

To begin with, the accuracy, which is the metric used in our experiments has the potential to overestimate the correctness of responses, even if the response does not accurately convey the intended semantic meaning. 
This discrepancy can occur because the metric simply verifies the existence of the answer entity, regardless of whether it is contextually appropriate. 
Unlike semantic parsing KGQA, which involves retrieving entities from the KG, or multiple-choice KGQA, where the answer is chosen from several options, evaluating metrics for generative KGQA remains an open field that warrants further investigation.

Secondly, there are cases where the tendency of LLMs to favor a certain fact verbalization method becomes overwhelming and difficult to manage.
As demonstrated in Tables 2 and 3, with the example of \llamachat on the \webqsp dataset, it has been observed that certain models may have a inclination for specific knowledge formats in particular datasets. 
Consequently, while our summarizer generally demonstrates good performance, controlling performance may become challenging when a specific model has a strong inclination towards a particular knowledge format.

Lastly, it is important to note that the performance of our proposed summarizer can be influenced by the performance of the retriever. 
As can be seen in Table \ref{tbl:retriever}, using a better retriever can lead to higher performance. 
While the off-the-shelf model we used (i.e. MPNet) demonstrates retrieving capabilities based on the semantic similarity between questions and facts, it's difficult to assert that it is a flawless retriever.
For example, in a 2-hop dataset(i.e. Mintaka), it tends to retrieve 1-hop neighbors more than 2-hops even if it is irrelevant. This is because an answer entity in 2-hop neighbors is unseen in given question, so that a retrieval model may measure question entities in 1-hop neighbors more similar.
Therefore, we are currently conducting further research to propose not only a more powerful summarizer but also a more flawless retrieving method simultaneously.

\section{Ethical Consideration}
\label{sec:ethical}
Throughout our research, we thoroughly explore our methodology using an open-source dataset, chosen to ensure transparency and integrity in our work.
It is important to acknowledge the inherent potential for biases within our summarizer, which relies on LLMs, and which may inadvertently reflect prevailing social biases. 
It is crucial to notice that our method is not intended to inflict harm upon any individuals or groups.

\section{Acknowledgements}
\label{sec:acknowledgements}
This work was supported by the IITP grants funded by the Korea government (MSIT) (No. RS-2020-II201361; RS-2024-00457882, AI Research Hub Project), and the NRF grant funded by the Korea government (MSIT) (No. RS-2023-00244689).


\bibliography{custom}
\bibliographystyle{acl_natbib}

\appendix

\section{Implementation Details}
\subsection{LLMs for Zero-Shot QA}
\label{subsec:llmdetail}

\begin{itemize}
    \item \textbf{Flan-T5} is a variant model grounded in the T5 architecture. T5, an encoder-decoder model, is trained on a text-to-text dataset featuring a diverse array of objectives. 
Flan-T5 represents an evolution of T5 through the process of instruction tuning, enhancing its performance by aligning it more closely with specific instructional contexts. 
We employ Flan-T5-XL for our QA model.

\item
\textbf{Llama2-7B-Chat} Llama2 is the developed version of Llama1\cite{touvron2023llama}. For the purpose of achieving more accurate measurement of summarizer performance, we employ the Llama2-7B-chat, which is specifically optimized for conversational contexts.

\item
\textbf{GPT-3.5-turbo}, developed by OpenAI, stands as a prominent closed-source model renowned for its high performance, making it particularly well-suited for application in QA models. 
Due to GPT-3.5-turbo being closed-source, we utilize the API provided by OpenAI\footnote{https://openai.com/api/}.
\end{itemize}

\subsection{Faithfulness Evaluation}
\label{subsec:hallucination}
To achieve a more faithful summary, we undertake distillation using data meticulously purged of summaries affected by hallucinations. 
For this purpose, we eliminate hallucinations from both broad and specific summaries based on prompts provided by G-eval~\cite{gralinski-etal-2019-geval}. 
We revise the G-eval-provided prompts with QA-specific instructions and generated auto Chain of Thought (CoT). 
This auto-generated CoT are then utilized as evaluation criteria to ascertain the factual consistency between the generated summaries and the facts, marking them as true or false. 
For a more accurate assessment of faithfulness, we measure hallucination using GPT-4, a more advanced model than GPT-3.5-turbo, which was used for reference summary generation and paraphrasing.
The faithfulness of the method is defined as (1 -hallucination occurrence rate).

\subsection{Reference Summary Generation}
For the purpose of generating reference summaries conducive to distillation, we utilize GPT-3.5-turbo. 
Given the scarcity of training data available within each dataset, we embark on a strategy of data augmentation to distill a more diverse array of summary cases. 
Through GPT-3.5-turbo, five summaries are augmented for each sample, and to foster an even greater diversity of summary instances, we adjust the decoding temperature, to 1.1.

\subsection{Paraphrased Summary Generation}
For generating specific summaries for DPO training, we conduct paraphrasing. 
Summary from the reference summaries that passed through the helpfulness and faithfulness filters is transformed from broader summary to specific summary through paraphrasing. 
For this paraphrasing endeavor, GPT-3.5-turbo is once again employed. 
To select the most effective summary among a variety of paraphrased summaries, multiple paraphrased summaries are generated for each sample. 
To ensure diversity, the decoding temperature is set to 1.1.

\subsection{Distilled Summary Generation}
In the generation of the final summary through the \proposed, the temperature is set to 0.1. 
While it is appropriate to elevate the temperature to consider a wider range of candidates during the creation of a augmented dataset for distillation, we conclude that for generating summaries for QA inference, it is more important to promote consistency over diversity in the summaries.

\section{Experimental Settings}
\subsection{Datasets}
\begin{itemize}

\item \textbf{WebQuestionsSP (WebQSP)} \cite{berant-etal-2013-semantic, yih-etal-2016-value} Semantic Parses Dataset, abbreviated to WebQSP, is a KGQA dataset providing SPARQL queries which allows a direct retrieval from Freebase KG. Due to the cessation of updates to Freebase we adopt Wikidata as the foundational KG. As we mentioned in section ~\ref{subsec:expset}, we utilized WebQSP-WD~\cite{sorokin-gurevych-2018-modeling}, a dataset that offers questions from WebQSP pre-linked to the Wikidata KG.    
\item \textbf{Mintaka}~\cite{sen-etal-2022-mintaka} \hspace{5pt} A KGQA dataset collected from Wikidata, with 8 complexity types, including `Count', `Comparative', `Superlative', `Ordinal', `Multi-hop', `Intersection', `Difference', `Yes/No', `Generic'. Question-Answer pairs are collected from Wikidata entities. This dataset is multilingual, and we use English datasets in this work.
\end{itemize}

\subsection{Various Retrievers}

\begin{itemize}
    \item \textbf{Random Knowledge} is a knowledge augmentation method that entails the selection of $K$ arbitrary facts to serve as the final knowledge.
    \item \textbf {Popular Knowledge} employs a method wherein triples are organized based on the releation frequency of their occurrence among one-hop neighbor triples. Then top K triples, sorted by the prevalence of their relations, are utilized as the final knowledge.
    \item \textbf{MPNet} is a retrieval method predicated on the semantic similarity between questions and triples. 
    Each triple in the triple set is compared to the question based on the cosine similarity between their MPNet representations.
    The K triples exhibiting the highest similarity to the question are then selected as the final knowledge.
\end{itemize}

\section{Additional Experiment Results}
\subsection{Ablation Study}

\begin{table}[t]
    \centering
    \begin{center}
    \small 
    \renewcommand{\arraystretch}{1.05}
    \begin{tabular}[h]{ll|c}
    \toprule
    \textbf{Dataset} & \textbf{Method} & \textbf{Acc} \\
    \midrule
    \midrule
    \multirow{4}{*}{\centering \textbf{WebQSP}} 
    & \textbf{{\textsc{EFSum}\textsubscript{$distill$}\xspace}} & \textbf{0.500} \\
    & \;\;\;\;w/o Paraphrase & \underline{0.477} \\ 
    & \;\;\;\;w/o Helpfulness & 0.453 \\ 
    & \;\;\;\;w/o All Filters, Parapharase & 0.469 \\ 
    \midrule
    \multirow{4}{*}{\centering \textbf{Mintaka}} 
    & \textbf{{\textsc{EFSum}\textsubscript{$distill$}\xspace}} & \textbf{0.338} \\
    & \;\;\;\;w/o Paraphrase & 0.296 \\ 
    & \;\;\;\;w/o Helpfulness & 0.289 \\ 
    & \;\;\;\;w/o All Filters, Parapharase & \underline{0.299} \\ 
    \bottomrule
    \end{tabular}
    \end{center}
    \vspace{-0.15in}
    \caption{\small Ablation study on different filters and paraphrasing. The best and second-best results are in \textbf{bold} and \underline{underlined}, respectively.}
    \label{tbl:ablation}
    \vspace{-0.1in}
\end{table}

To investigate the effect of different filters and paraphrasing approach on \proposed, in Table \ref{tbl:ablation}, we evaluate the performance of \proposed on ablation study. 
For the experimental setting, we used Flan-T5-XL as a QA model on WebQSP and Mintaka with a maximum token length of contextual knowledge as $L=400$. 
Each row indicates whether each filter or paraphrasing approach is used or not. 
\textquoteleft w/o All Filters, Paraphrase\textquoteright means the model is trained without using any filtering or paraphrasing technique.
Table \ref{tbl:ablation} demonstrates that the performance of the distilled model drops filters are removed or the paraphrasing stage is omitted. 
This result indicates that \proposed shows the best performance when all processes are put together. 
When the helpfulness filter is removed, some summaries that do not aid the QA task get mixed into the chosen set during DPO training. This disrupts the model's optimization process, leading to a performance decline. 
Consequently, removing the helpfulness filter results in the most critical performance drop. 
The paraphrasing stage also plays a crucial role. Without paraphrasing, the ability to highlight important information within paragraphs decreases. 
As a result, key information in the summary becomes distracted, making it difficult for the QA model to digest the knowledge. 
This inevitably leads to a performance decline.
\subsection{Generalization Ability}
\begin{table}[t]
    \centering
    \begin{center}
    \resizebox{0.475\textwidth}{!}{
    \renewcommand{\arraystretch}{1.05}
    \begin{tabular}[h]{ccc|c} 
    \toprule
    \textbf{Method} & \textbf{Trained on} & \textbf{Generate Summaries on} & \textbf{Acc} \\
    \midrule 
    \midrule
    \textsc{EFSum}\textsubscript{$distill$}\xspace & Mintaka & WebQSP & 0.455 \\
    \midrule
    \textsc{EFSum}\textsubscript{$distill$}\xspace & WebQSP & Mintaka & 0.281 \\
    \midrule
    KAPING & WebQSP & WebQSP & 0.439 \\
    \midrule
    Rewrite & Mintaka & Mintaka & 0.288 \\
    \bottomrule
    \end{tabular}
    }
    \end{center}
    \vspace{-0.15in}
    \caption{\small Generalization capability of \textsc{EFSum}\textsubscript{$distill$}\xspace for unseen datasets.}
    \label{tbl:generalization}
    \vspace{-0.1in}
\end{table}

We examine the generalization capability of \proposed for unseen datasets. 
In Table ~\ref{tbl:generalization}, We investigate the cross-dataset experiments on Flan-T5-XL with a maximum token length of contextual knowledge as $L=400$ (i.e. compatible to Table ~\ref{tbl:main_density}). 
We use {\textsc{EFSum}\textsubscript{$distill$}\xspace} that generates the summary for Mintaka trained on WebQSP, also vice versa. 
Referring to Table \ref{tbl:main_density}, the most superior baselines scores 0.439 and 0.288 for WebQSP and Mintaka, respectively. 
And {\textsc{EFSum}\textsubscript{$distill$}\xspace} that trained on unseen dataset scores 0.455 and 0.281 for WebQSP and Mintaka, respectively.
This result shows that our methods outperform the most of baselines and are competitive to the most effective baselines. 
This indicates that the proposed evidence-focused approach can effectively  summarize the evidence even on unseen datasets.

\section{Qualitative Examples}

\subsection{Case Study}
\begin{table*}[thbp]
    \centering
    \small
    \begin{tabular}{p{14cm}}
    
    \toprule
    \textbf{Question:}
    Who is Emilio Estevez's father?
    
    \textbf{Answer:}
    Martin Sheen
    \\\midrule
    
    \textbf{\kaping:}
     (Emilio Estevez's, family name, Estévez), (Emilio Estevez's, given name, Emilio), (Emilio Estevez's, father, Martin Sheen), (Emilio Estevez's, occupation, actor), (Joe Estevez, relative, Emilio Estevez), (Ramón Estévez, sibling, Emilio Estevez), (Emilio Estevez's, relative, Joe Estevez), (Ramón Estévez, father, Martin Sheen), (Martin Sheen, child, Emilio Estevez), (Emilio Estevez's, date of birth, +1962-05-12T00:00:00Z), (Joe Estevez, relative, Ramón Estévez), (Emilio Estevez's, birth name, Emilio Estevez Sheen), (Emilio Estevez's, sibling, Ramón Estévez), (Ramón Estévez, relative, Joe Estevez), (Ramón Estévez, family name, Estévez), (Emilio Estevez's, occupation, film producer), (Renée Estévez, sibling, Emilio Estevez), (Charlie Sheen, sibling, Emilio Estevez), (Janet Sheen, child, Emilio Estevez), (Martin Sheen, child, Ramón Estévez), (Joe Estevez, family name, Estevez), (Martin Sheen, family name, Estévez), (Emilio Estevez's, described by source, Obalky knih.cz), (Ramón Estévez, occupation, television actor), (Emilio Estevez's, occupation, film director), (Emilio Estevez's, country of citizenship, United States of America), (Emilio Estevez's, occupation, writer), (Estévez, said to be the same as, Estevez), (Emilio Estevez's, sibling, Renée Estévez), (Emilio Estevez's, place of birth, New York City)\\\midrule

    \textbf{\kgtotext}
    the family name of emilio estevez is estévez . he was born in new york city and is from the united states . he is the father of two children , both of whom are children . he is the actor , writer and film director . he is the brother of charlie sheen and his sister is renée .\\\midrule
    
    \textbf{\rewrite:}
    Emilio Estevez's family name is Estévez. Emilio Estevez's given name is Emilio. Emilio Estevez's father is Martin Sheen. Emilio Estevez is an actor. Joe Estevez is a relative of Emilio Estevez. Ramón Estévez is a sibling of Emilio Estevez. Emilio Estevez is related to Joe Estevez as they are relatives. Ramón Estévez is the father of Martin Sheen. Martin Sheen is the father of Emilio Estevez. Joe Estevez is a relative of Ramón Estévez. Emilio Estevez has a sibling named Ramón Estévez. Ramón Estévez is a relative of Joe Estevez. Ramón Estévez's family name is Estévez. Emilio Estevez is a film producer. Renée Estévez is the sibling of Emilio Estevez. Charlie Sheen is the sibling of Emilio Estevez. Janet Sheen is the mother of Emilio Estevez. Martin Sheen's child is Ramón Estévez. Joe Estevez's family name is Estevez. Martin Sheen's family name is Estévez. Emilio Estevez is described by the source Obalky knih.cz. Ramón Estévez is a television actor. Emilio Estevez is a film director. Emilio Estevez is a citizen of the United States of America. Emilio Estevez is a writer. Estévez is said to be the same as Estevez. Emilio Estevez has a sibling named Renée Estévez. Emilio Estevez was born in New York City. Ramón Estévez is an actor. Charlie Sheen's family name is Estévez. \\\midrule
    
    \textbf{\proposed}
    Emilio Estevez's father is Martin Sheen. He was born in New York City in 1962 and is known for his work as an actor, film director, and writer. He has siblings named Ramón Estévez, Joe Estevez, Renée Estévez, and Charlie Sheen. Martin Sheen is the father of both Emilio Estevez and Ramón Estévez.\\
    \bottomrule
    \end{tabular}
    \caption{An example of verbalized facts for Mintaka.}
    \label{tbl:casestudy}
\end{table*}
In Table~\ref{tbl:casestudy}, we present an example of verbalized facts generated by each method.
\proposed emphasizes evidence necessary to answer the question (i.e., \textit{Emilio Estevez’s father is Martin Sheen.}) at the beginning, while excluding irrelevant details for brevity, as opposed to other baselines. 
Duplicated relations (i.e., relations of \textit{siblings}) are aggregated, and it is clear that our method is more compact than \kaping and \rewrite. Comparing to \kgtotext, it also sums up the given triples, but reduces a loss of essential information.

\subsection{Prompts}
\label{subsec:detailedprompt}
We provide various LLM prompts, used for (1) evidence-focused summarization, (2) knowledge-augmented zero-shot QA, (3) paraphrase of summary candidates, and (4) faithfulness evaluation.
\begin{table*}[thbp]
    \small
    \centering
    \begin{tabular}{p{14cm}}
    \toprule
    \textbf{Evidence-Focused Summarization Prompt (\proposedprompt)} \\
    \midrule
\textcolor{teal}{\textbf{[Task Description]}}\\
You are a knowledge graph summarizer for Question Answering. I will give you ``Question'', ``Fact triples''. You should turn triples into *summary*. The *summary* should serve as a context to facilitate QA (Question and Answer) tasks. \\
\textbf{\#\# Caution1:} The *summary* should not explicitly mention what the correct answer is. \\
\textbf{\#\# Caution2:} The *summary* should only contain information of the given triples. \\
\textbf{\#\# Caution3:} Each triplet is seperated with ``\textbackslash n'' and head, relation, tail are provided in head | relation | tail format. \\
\textbf{\#\# Question:}  \{\} \\
\textbf{\#\# Fact triples:} \{\} \\
\textbf{\#\# Summary:}
        \\ \bottomrule
    \end{tabular}
    \caption{The prompt for evidence-focused summarization.}
    \label{tbl:sum_prompt}
\end{table*}

\begin{table*}[thbp]
    \small
    \centering
    \begin{tabular}{p{14cm}}
    \toprule
    \textbf{KG-Augmented Question Answering Prompt (\kaping)} \\
    \midrule
\textcolor{teal}{\textbf{[Task Description]}}\\
You are a student who have to solve the question. I’ll give you a triples as a context. But if it is not useful, just ignore it and generate your own guess. \\
\textbf{\#\# Triples:} \{\} \\
\textbf{\#\# Question:} \{\} \\
\textbf{\#\# You are aware of the answer. Generate only short answer(You have to guess something):}
    \\\bottomrule
    \\
    \toprule
    \textbf{KG-Augmented Question Answering Prompt (\proposed)} \\
    \midrule
\textcolor{teal}{\textbf{[Task Description]}}\\
You are a student who have to solve the question. I’ll give you a summary as a context. But if it is not useful, just ignore it and generate your own guess. \\
\textbf{\#\# Summary: \{\}} \\
\textbf{\#\# Question: \{\}} \\
\textbf{\#\# You are aware of the answer. Generate only short answer(You have to guess something): }
    \\\bottomrule
    \end{tabular}
    \caption{The prompts for knowledge-augmented zero-shot question answering.}
    \label{tbl:qa_prompt}
\end{table*}

\begin{table*}[thbp]
    \small
    \centering
    \begin{tabular}{p{14cm}}
    \toprule
    \textbf{Summary Candidate Paraphrasing Prompt} (In case that answer does not exist in the summary candidate) \\
    \midrule
\textcolor{teal}{\textbf{[Task Description]}}\\
You are a knowledge graph summarizer for Question Answering. I will give you ``Question'', ``Knowledge Summary''. You should pharaphrase the original ``Knowledge Summary''. The paraphrased summary should serve as a context to facilitate QA (Question and Answer) tasks. Paraphrase the original “Knowledge Summary” to be more helpful to solve the QA. \\
\textbf{\#\# Question:} \{\} \\
\textbf{\#\# Original Summary:} \{\} \\
\textbf{\#\# Paraphrased Summary:}
        \\ \bottomrule
        \\
            \toprule
    \textbf{Summary Candidate Paraphrasing Prompt} (In case that answer does exist in the summary candidate) \\
    \midrule
\textcolor{teal}{\textbf{[Task Description]}}\\
You are a knowledge graph summarizer for Question Answering. I will give you 
``Question'', ``Answer'', ``Knowledge Summary''. You should pharaphrase the original ``Knowledge Summary''. The paraphrased summary should serve as a context to facilitate QA (Question and Answer) tasks. Paraphrase the original ``Knowledge Summary'' to be more helpful to solve the QA. \\
\textbf{\#\# Question:} \{\} \\
\textbf{\#\# Answer:} \{\} \\
\textbf{\#\# Original Summary:} \{\} \\
\textbf{\#\# Paraphrased Summary:}
        \\ \bottomrule
    \end{tabular}
    \caption{The prompt for summary candidate paraphrasing.}
    \label{tbl:sum_prompt}
\end{table*}
\begin{table*}[thbp]
    \small
    \centering
    \begin{tabular}{p{14cm}}
    \toprule
    \textbf{Summary Candidate Faithfulness Evaluation Prompt} \\
    \midrule
\textcolor{teal}{\textbf{[Task Description]}}\\

You will be given one summary written to provide useful contexts by given source triples from knowledge graphs. Your task is to check whether the given summary induces factual inconsistency. Please make sure you read and understand these instructions carefully. Please keep this evaluation creteria open while reviewing, and refer to it as needed. \\
\textbf{Evaluation Criteria:} \\
Factual Inconsistency (0 or 1): Does the summary untruthful or misleading facts that are not supported by the source triples? If does, mark 1. Otherwise, mark 0. \\
\textbf{Evaluation Steps:} \\
1. read and understand the source triples first. note the entities that are in focus and the relations between them. \\
2. proceed to read through the summary provided. \\
3. compare the information in the summary with that in the source triples. pay particular attention to the entities, actions, and relations. \\
4. mark ``1'' if the summary contains factual inconsistencies, i.e., if it states untruthful or misleading facts that are not supported by the source triples. \\
5. mark ``0'' if the summary is consistent with the source triples and does not misrepresent the facts provided by the source triples. \\
Remember, you are not assessing the quality of the writing, but the factual consistency of the summary compared to the source triples. perfection in grammar or style does not account for factual consistency. conversely, poor grammar or style does not necessarily mean factual inconsistency. the key lies in the alignment of facts between the source triples and the summary. \\
\textbf{\#\# Source Triples:} \{\} \\
\textbf{\#\# Summary:} \{\} \\
\textbf{\#\# Does the summary contain factual inconsistency?} Answer:
        \\ \bottomrule
    \end{tabular}
    \caption{The prompt for faithfulness evaluation.}
    \label{tbl:sum_prompt}
\end{table*}

\end{document}